\documentclass{article}

    \PassOptionsToPackage{numbers, compress}{natbib}
\usepackage{neurips_2020}

\usepackage[utf8]{inputenc} % allow utf-8 input
\usepackage[T1]{fontenc}    % use 8-bit T1 fonts
\usepackage{hyperref}       % hyperlinks
\usepackage{url}            % simple URL typesetting
\usepackage{booktabs}       % professional-quality tables
\usepackage{amsfonts}       % blackboard math symbols
\usepackage{nicefrac}       % compact symbols for 1/2, etc.
\usepackage{microtype}      % microtypography
\usepackage{graphicx}
\usepackage{float}
\usepackage[english]{babel}

\usepackage{natbib}
\title{Spoiler Alert: Using Natural Language Processing to Detect Spoilers in Book Reviews}

\author{%
  Marshall Ho\thanks \\
  Department of Computer Science\\
  University of Toronto\\
  \texttt{marshall.ho@utoronto.ca} \\
}

\begin{document}

\maketitle

\begin{abstract}
This paper presents an NLP (Natural Language Processing) approach to detecting spoilers in book reviews, using the University of California San Diego (UCSD) Goodreads Spoiler dataset. We explored the use of LSTM, BERT, and RoBERTa language models to perform spoiler detection at the sentence-level. This was contrasted with a UCSD paper which performed the same task, but using handcrafted features in its data preparation. Despite eschewing the use of handcrafted features, our results from the LSTM model were able to slightly exceed the UCSD team’s performance in spoiler detection.

\end{abstract}

\section{Introduction}

In this report, we will explore the task of spoiler detection using the UCSD Goodreads Spoiler dataset [\citenum{wan_fine-grained_2019}]. A spoiler is a piece of information in a movie or book review which reveals important plot elements, such as the ending or a major plot twist. Our models are designed to flag spoiler sentences automatically.

\subsection{Related Work}
There is little existing work done on spoiler detection. SpoilerNet, a custom network described in \emph{Fine-grained spoiler detection from large-scale review corpora} from UCSD, is one of the leading models in spoiler detection [\citenum{wan_fine-grained_2019}]. SpoilerNet is a bi-directional attention based network which features a word encoder at the input, a word attention layer and finally a sentence encoder. It achieved an impressive AUC (area under the curve) score of 0.889, indicating a relatively high true positive:false positive ratio. 

Wan et al. introduced a handcrafted feature: DF-IIF - Document Frequency, Inverse Item Frequency - to provide their model with a clue of how specific a word is. This would allow them to detect words that reveal specific plot information.

\begin{figure}[h]
\includegraphics[scale=0.5]{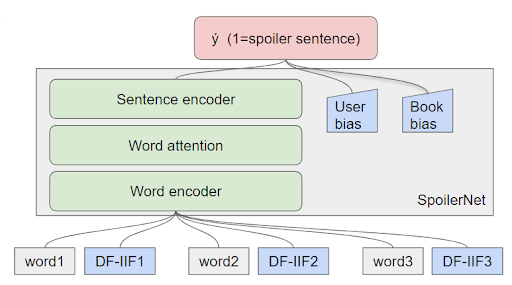}
\centering
\caption{UCSD SpoilerNet}
\end{figure}
\subsection{Novelty}

Our models rely less on handcrafted features compared to the UCSD team. We make use of an LSTM model and two pre-trained language models, BERT and RoBERTa, and hypothesize that we can have our models learn these handcrafted features themselves, relying primarily on the composition and structure of each individual sentence. Through these methods, our models could match, or even exceed the performance of the UCSD team.

\section{Data Exploration and Preprocessing}
\subsection{Dataset}
The UCSD Goodreads Spoilers dataset was created as part of the UCSD Book Graph project [\citenum{noauthor_ucsd_nodate}]. The UCSD team scraped more than 15M records with detailed reviews on books from goodreads.com. This dataset is very skewed - only about 3\% of review sentences contain spoilers. It is also context-sensitive: spoiler flagging depends heavily upon the book being reviewed. For example, “the main character died” spoils “Harry Potter” far more than the Bible.

\subsection{Exploratory Analysis}
The Goodreads Spoiler dataset has 7 columns: book\_id, user\_id, review\_id, rating, has\_spoiler, review\_sentences, and timestamp. The most important feature is “review\_sentences”, a list of all sentences in a particular review. Each sentence in the list is preceded by its own “spoiler” flag -- “1” for spoiler, “0” otherwise. Spoiler tags are self-reported by the review authors [\citenum{wan_fine-grained_2019}]. The dataset has about 1.3 million reviews.  We extracted a subset of 275,607 reviews randomly and performed an exploratory analysis. One finding was that spoiler sentences were typically longer in character count, perhaps due to containing more plot information, and that this could be an interpretable parameter by our NLP models.

\begin{table}[H]
\begin{center}
\caption{Exploratory data from the Goodreads Spoiler dataset}

    \begin{tabular}{|l|l|}
\hline
Unique users                            & 17405 (6.3\%)              \\ \hline
Reviews per user                        & 15.8 (mean), 8 (median)    \\ \hline
Unique books                            & 24873 (9.0\%)              \\ \hline
Reviews per book                        & 11.1 (mean), 5 (median)    \\ \hline
Total sentences                         & 3,534,334                  \\ \hline
No-spoiler sentences                    & 3,419,580 (96.8\%)         \\ \hline
Length of spoiler sentences (chars)     & 88.1 (mean), 74.0 (median) \\ \hline
Length of non-spoiler sentences (chars) & 81.4 (mean), 70.0 (median) \\ \hline
\end{tabular}
\end{center}
\end{table}

\subsection{Dataset preparation}
Our final dataset consists of the review\_sentences feature, flattened out so that each sample would contain one review sentence. Spoiler flags were separated into a separate label array corresponding to each sentence. We also explored other related UCSD Goodreads datasets, and decided that including each book’s title as a second feature could help each model learn the more human-like behaviour, having some basic context for the book ahead of time. This was important as the original dataset’s only information about the book itself was through a non-descriptive book\_id feature. For each of our models, the final size of the dataset used was approximately 270,000 samples in the training set, and 15,000 samples in the validation and test sets each (used for validating results).

\section{Models}
We found both LSTM [\citenum{hochreiter_long_1997}] and BERT [\citenum{devlin_bert_2019}] to be capable candidates due to their success with sentiment analysis in part 1 of our projects. We also aspired to include a model newer than BERT, and decided to go with the better optimized RoBERTa [\citenum{liu_roberta_2019}], after considering XLNet [\citenum{yang_xlnet_2020}] as well.

The predictive task for our models is to flag each spoiler sentence individually. We will compare our work with that of Wan et al. at UCSD. Since the spoiler data is heavily skewed, AUC was used as a metric. The UCSD team achieved a score of 0.889 [\citenum{wan_fine-grained_2019}] and that will be used as our baseline. 

\subsection{Model 1: LSTM}
Long short-term memory (LSTM) is a recurrent neural network (RNN) which excels in processing sequences [\citenum{hochreiter_long_1997}]. A LSTM cell passes on a hidden state over time steps and the input, output and forget gates regulate the flow of information into and out of the cell [\citenum{hochreiter_long_1997}]. The LSTM’s major shortcoming is its size and complexity, taking a substantial amount of time to run compared with other methods. However, the nature of the input sequences as appended text features in a sentence (sequence) makes LSTM an excellent choice for the task.

Each text entry was put through the Keras Tokenizer function, which maps each unique vocabulary word to a numeric entry before vectorizing each sentence to become compatible with the LSTM model. Hyperparameters for the model included the maximum review length (600 characters, with shorter reviews being padded to 600), total vocabulary size (8000 words), two LSTM layers containing 32 units, a dropout layer to handle overfitting by inputting blank inputs at a rate of 0.4, and the Adam optimizer with a learning rate of 0.003. The loss used was binary cross-entropy for the binary classification task.

\subsection{Model 2: BERT}
In sentiment analysis tasks (seen in part 1 of our course project) pre-trained language models such as BERT outperformed LSTMs. The attention-based nature of BERT means entire sentences can be trained simultaneously, instead of having to iterate through time-steps as in LSTMs. The bi-directional nature of BERT also adds to its learning ability, because the “context” of a word can now come from both before and after an input word.

We chose to use Hugging Face’s “bert-base-cased” model, which has 12 layers, and 109M parameters, producing 768-dimensional embeddings with a model size of over 400MB [\citenum{noauthor_bert_nodate}]. We fed the same input -- concatenated “book title” and “review sentence” -- into BERT. We used a dropout layer and then a single output neuron to perform binary classification. We did not use sigmoid activation for the output layer, as we chose to use BCEWithLogitsLoss as our loss function which is quicker and provides more mathematical stability.

\subsection{Model 3: RoBERTa}
We also wanted to try out a third model, and researched several candidates. We chose RoBERTa, which is a better-optimized BERT model developed by Facebook’s AI team [\citenum{noauthor_bert_nodate}]. We used “roberta-base” from Hugging Face [\citenum{noauthor_roberta_nodate}], which has 12 layers and 125 million parameters, producing 768-dimensional embeddings with a model size of about 500MB. The setup of this model is similar to that of BERT above.

\section{Experiments and Results}
We trained our 3 networks with the flattened dataset of 1 million sentences. Metrics for true/false positives/negatives are collected in order to calculate AUC. We typically ran 5 epochs per training session. The first versions of our models trained on the review sentences only (without book titles); the results were quite far from the UCSD AUC score of 0.889. Follow-up trials were performed after tuning hyperparameters such as batch size, learning rate, and number of epochs, but none of these led to substantial changes. After studying supplementary datasets related to the UCSD Book Graph project (as described in section 2.3), another preprocessing data optimization method was found. Including book titles in the dataset alongside the review sentence could provide each model with additional context. For instance, the model could capture the presence of shared words between each title and review sentence (ex. The word “Harry” appearing in a spoiler-sentence, given the title “Harry Potter”). The highest AUC score we achieved was approximately 0.91, from the LSTM model.

\begin{figure}[t]
\includegraphics[scale=0.6]{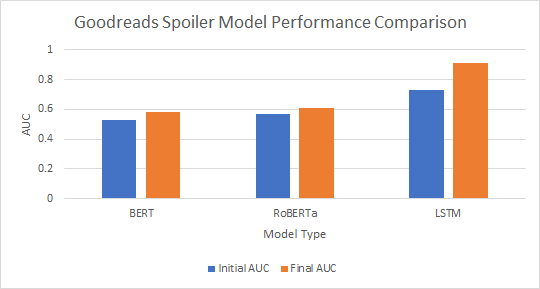}
\centering
\caption{Initial vs final model AUC results.}
\end{figure}

Each of our 3 team members maintained his own code base. We used a mix of development environments (Google Colab, vscode) and platforms (TensorFlow, PyTorch). Our BERT and RoBERTa models have subpar performance, both having AUC close to 0.5. LSTM was much more promising, and so this became our model of choice.

\section{Discussion}
The AUC score of our LSTM model exceeded the lower end result of the original UCSD paper. While we had been confident with our innovation of adding book titles to the input data, beating the original work in such a short period of time exceeded any reasonable expectation we had. This implies that spoiler detection does not require handcrafted features such as DF-IIF, and that text sequences are mostly sufficient for predicting labels with a relatively abstract relationship with the text. Supplemental context (titles) help boost this accuracy even further.

Our LSTM performed substantially better than our BERT and RoBERTa models, for which the AUC scores placed solidly in the 0.50’s. We are hesitant in drawing bold conclusions, as we would have spent more time debugging various parts of the model and the metric collection code, given more time. One possible argument is that BERT, from which RoBERTa is derived, is not pre-trained to handle a highly skewed dataset (only ~3\% of sentences contained spoilers). A model trained specifically for sentiment analysis or text classification on more balanced data, such as BERT, may not have been the perfect candidate for spotting the intricacies of this dataset.

The definition of a spoiler is inherently subjective much of the time - and since this dataset was human-labelled, it is certainly subject to mislabeling. Thankfully, the sheer number of samples likely dilutes this effect, but the extent to which this occurs is unknown. 

\subsection{Future Work}
The first priority for the future is to get the performance of our BERT and RoBERTa models to an acceptable level. For the scope of this investigation, our efforts leaned towards the winning LSTM model, but we believe that the BERT models could perform well with proper adjustments as well. We may also test the inclusion of more features such as user information into the input of our models. We are also looking forward to sharing our findings with the UCSD team.

\section*{Attributions and Code}
Each member of our team contributed equally. Allen Bao created our LSTM model and created our first dataset. He also optimized his network to match and exceed UCSD performance. Marshall Ho studied the problem of our low initial AUC scores and created the second dataset which added book titles. He also developed our model based on RoBERTa. Saarthak Sangamnerkar developed our BERT model and the metric reporting framework in pytorch. He is also our LaTeX expert.

Code repository: https://github.com/allenbao64/csc2515-project
\medskip
\bibliography{references}

\end{document}